\documentclass[10pt,letterpaper]{article}

\usepackage{hyperref}
\usepackage{cogsci}
\usepackage{pslatex}
\usepackage{apacite}
\usepackage{graphicx}
\usepackage{xcolor}
\usepackage{amsmath}
\usepackage{gensymb}

\title{Multifunctionality in embodied agents: Three levels of neural reuse}

\author{{\large \bf Madhavun Candadai (madcanda@indiana.edu)} \\
  Cognitive Science Program, Indiana University \\
  Bloomington, IN 47406 USA
  \AND {\large \bf Eduardo J. Izquierdo (edizquie@indiana.edu)} \\
  Cognitive Science Program and School of Informatics, Computing, and Engineering, Indiana University\\
  Bloomington, IN 47406 USA}

\begin{document}

\maketitle

\begin{abstract}
The brain in conjunction with the body is able to adapt to new environments and perform multiple behaviors through reuse of neural resources and transfer of existing behavioral traits. Although mechanisms that underlie this ability are not well understood, they are largely attributed to neuromodulation. In this work, we demonstrate that an agent can be multifunctional using the same sensory and motor systems across behaviors, in the absence of modulatory mechanisms. Further, we lay out the different levels at which neural reuse can occur through a dynamical filtering of the brain-body-environment system's operation: structural network, autonomous dynamics, and transient dynamics. Notably, transient dynamics reuse could only be explained by studying the brain-body-environment system as a whole and not just the brain. The multifunctional agent we present here demonstrates neural reuse at all three levels.

\end{abstract}

\section{Introduction}
A crucial aspect to adaptation in cognitive beings is their ability to exploit regularities in the environment and reuse existing resources across multiple behaviors. Extensive empirical evidence shows that neural resources optimized during the course of learning one behavior are reused for others~\cite{Anderson:2010}. This multi-functional ability of neural circuits has been demonstrated in the small nervous systems of the nematode worm {\it Caenorhabditis elegans} (302 neurons)~\cite{Hobert:2003} as well as in the macro scale of the human brain (100 billion neurons)~\shortcite{Lizier:2011}. The mechanisms that facilitate this phenomenon have largely been attributed to neuromodulation and synaptic plasticity~\cite{Briggman:2008,Getting:1989,Morton:1994}.

\begin{figure}[!t]
    \centering
    \includegraphics[width=3.2in]{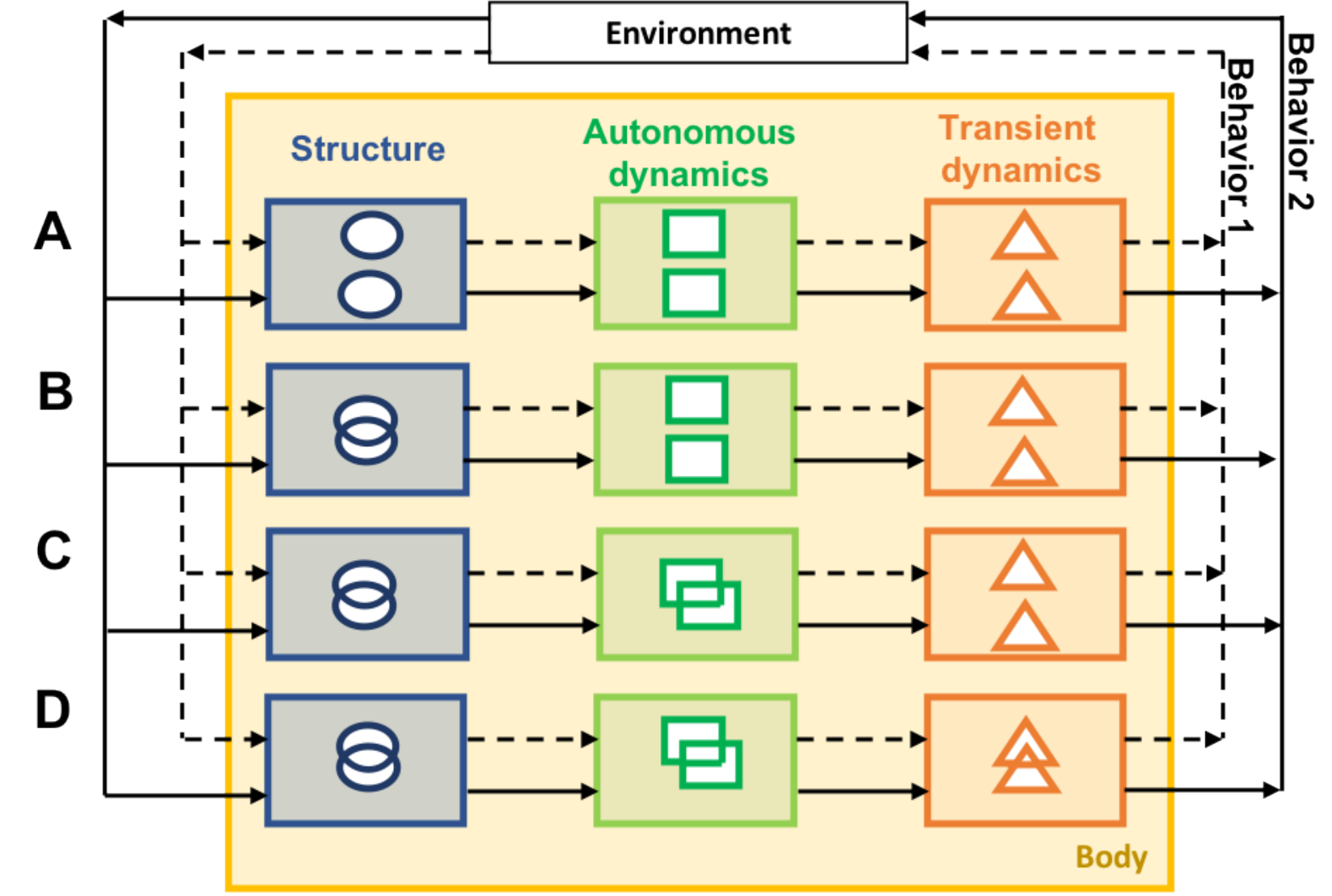}
    \caption{Three levels of neural reuse in multifunctional agents: structure (blue), autonomous dynamics (green), and transient dynamics (orange).
    [A] Dedicated circuits for each behavior: non-overlapping structures, and thus non-overlapping autonomous or transient dynamics.
    [B-D] Multifunctional circuits: overlapping structures
    (i.e., shared neural resources for multiple behaviors).
    [B] Autonomous dynamics are unique to each behavior, and thus transient dynamics are also unique
    (i.e., different set of attractors for each behavior).
    [C] Autonomous dynamics are shared across multiple behaviors, but transient dynamics are unique to each behavior
    (i.e., overlapping set of attractors for multiple behaviors, but different dynamics when coupled with the body and environment).
    [D] Both autonomous and transient dynamics are shared across multiple behaviors
    (i.e., overlapping set of attractors and similar overall dynamics when coupled with the body and environment for multiple behaviors).
    }
    \label{fig:schematic}
\end{figure}

The goal of this work is to show a concrete example of how the interaction between brain, body and environment enables neural networks to perform multiple behaviors and elucidate the dynamical aspects that lead to it. Reuse in embodied recurrent neural networks unfold over three levels: structural network, autonomous dynamics of the neural network, and transient dynamics of the neural network (Fig.~\ref{fig:schematic}).
Structure is defined by the neural circuit itself, the intrinsic parameters of the neurons, and the synaptic strength of connectivity between them. While it is possible that an agent possesses specialized circuits for performing different behaviors (Fig.~\ref{fig:schematic}A), reuse at this level involves utilizing overlapping circuits to produce multiple behaviors (Fig.~\ref{fig:schematic}B,C,D).
The next level, when structure is reused, is that of the neural network's autonomous dynamics isolated from the body. Each behavior is associated with a set of phase-portraits corresponding to the inputs the agent experiences while performing them. The sets of phase-portraits (and the attractors therein) could be overlapping (Fig.~\ref{fig:schematic}C,D) or could be unique to each behavior (Fig.~\ref{fig:schematic}B). The set of all attractors from all phase-portraits corresponding to a behavior are also referred to as the attractor set of the behavior in this paper.
The third level of reuse is that of ongoing transient dynamics as the agent is in continuous closed-loop interaction with the environment. When there is attractor reuse from the previous level, it is possible that multiple behaviors navigate different transients around those attractors (Fig.~\ref{fig:schematic}C) or, they might be reused too (Fig.~\ref{fig:schematic}D).

We show, for the first time to our knowledge, that reuse of transients, namely indistinguishable neural activity, can produce starkly distinct behaviors in embodied dynamical neural networks. We also show that this can only be observed when the brain is studied in conjunction with the closed-loop interaction between the body and the environment. Specifically, we optimized embodied agents with dynamical neural controllers to perform two tasks (object categorization and pole-balancing). We then analyzed the best agent to reveal how neural resources are reused across the three aforementioned levels to perform the two tasks.

\section{Methods}
\label{sec:methods}

\subsection{Agent}
\begin{figure}
    \centering
    \includegraphics[width=0.4\textwidth]{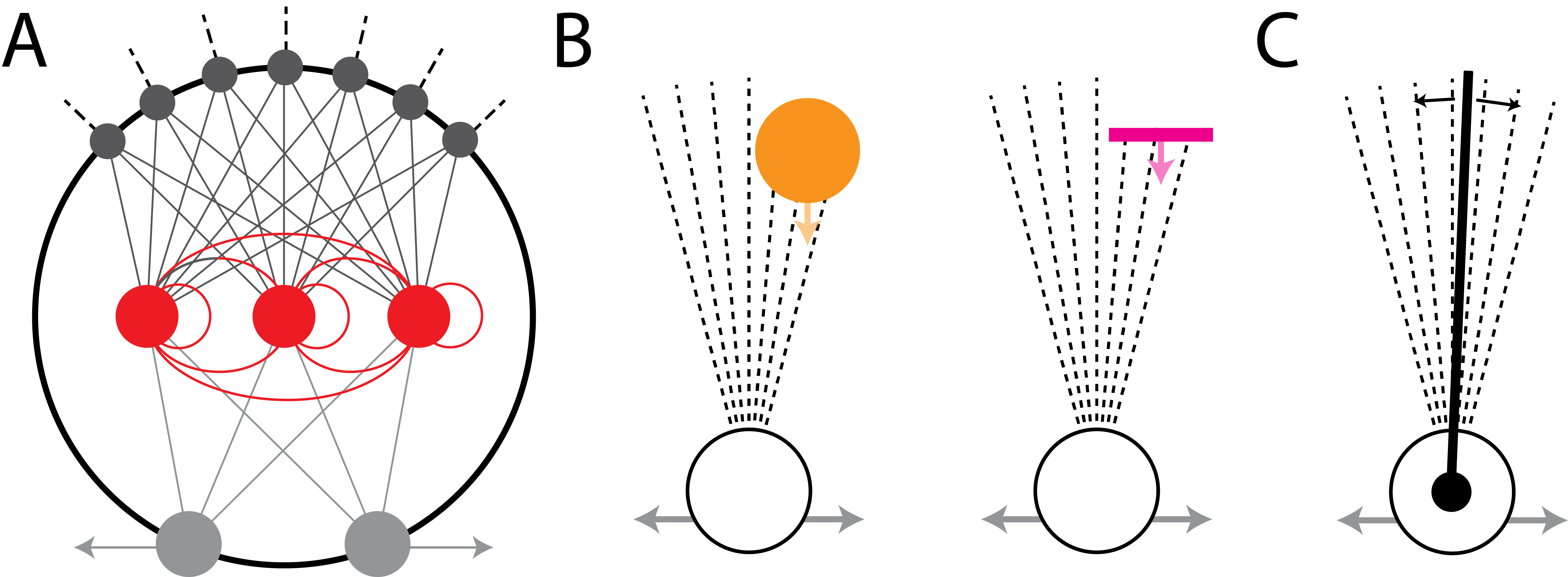}
    \caption{Agent design and task setup.
    [A]~7 rays of vision feed into sensory neurons (black). These neurons are fully connected to the recurrent interneuron layer (red), that in turn feed the left and right motor neurons (grey).
    [B]~Categorization task with circle and line trial. The falling object needs to be caught if it is a circle and avoided if it is a line.
    [C]~Pole-balancing task. The pole attached to the agent's center is expected it to keep balanced within the rays.}
    \label{fig:setup}
\end{figure}

The agent design is identical to that introduced in earlier work~\cite{Beer:1996}. The agent is circular with a diameter of 30 units and is equipped with 7 sensory rays radiating from its center, equally distributed over an angle of $\pi/6$. The rays constitute the ``eye'' of the agent and have a range of 265 units. Each sensory ray feeds into a sensory neuron, and the magnitude of input is inversely proportional to the distance at which that ray is intersected by an object (Fig.~\ref{fig:setup}). The sensory neurons are stateful units that are governed by the following dynamics $\tau_s\dot{s}_i = -s_i + I_i$ where $i \in [1,7]$, $\tau_s$ is the time-constant, fixed to be the same across all 7 neurons, $s_i$ is the state of the neuron, and $I_i$ is the sensory input received from the corresponding ray. Neuron output is $o_i = \sigma(-g_s(s_i+\theta_s))$ where $\sigma(x)=1/(1+exp(-x))$ is the sigmoid function of the state with a gain, $g_s$, and bias $\theta_s$, that are also fixed to be the same across all sensory neurons. The sensory neurons are fully connected to an interneuron layer, made up of a fully recurrently connected continuous-time recurrent neural network (CTRNN). The interneurons are governed by

\begin{equation}
    \tau_i\dot{s}_i = -s_i + \sum_{j=1}^{N}{w_{ji}\sigma(g_j(s_j+\theta_j))} + \sum_{k=1}^{7}{w_{ki}o_k}\\
\end{equation}

\noindent where $\tau_i$ is the time-constant of interneuron $i$, $s_i$ is its state, $N$ is the number of interneurons, $w_{ji}$ is the weight from neuron $j$ to neuron $i$, $g_j$ is the gain from incoming neuron $j$, $s_j$ is its state and $\theta_j$ its bias. The last term refers to the input to interneuron $i$ defined as the weighted sum of the outputs of the sensory neurons, $o_k$, with the weight from the $k^{th}$ sensory neuron being $w_{ki}$. The interneurons project to two motor neurons that take on dynamics similar to the interneurons. However, unlike the interneurons they are not recurrently connected. The motor neurons control the effective acceleration of the agent, a, as follows $a = g_m(\sigma(s_r+\theta_m) - \sigma(s_l+\theta_m))$ where $g_m$ refers to the gain, $s_r$ and $s_l$ are the internal state variables for the right and left motor neurons respectively, $\sigma$ is the standard sigmoidal activation function and $\theta_m$ is the common bias term for both motor neurons. The agent and environment are continuous time systems, simulated using Euler integration with a step size of 0.1.

\subsection{Categorization task}
We replicated the categorization task first introduced in Beer (1996). The task involves discriminating between the shape of falling objects (circles and lines) by moving towards one (circles) and away from the other (lines). The circles' diameter and the line's length were both set at 30 units. To encourage generalization, each evaluation of the agent's performance was conducted over 8 different trials for each type of object, with the objects' initial horizontal offset from the agent uniformly distributed in the range $[-50,50]$. The objects fall with a constant velocity of 0.3 units per second. Performance in this task was quantified by averaging over $1-|d_i|$ for the circle trials and $|d_i|$ for the line trials; where $d_i$ is the normalized distance between the center of the agent and the center of the object when the vertical offset between the agent and the object reaches 0 (offset of over 45 units was clipped at 45).

\subsection{Pole-balancing task}
We adapted the original pole-balancing task~\shortcite{Barto:1984} such that the agent has the pole attached to its center and senses it through the same rays used for sensing the falling objects~\cite{Vasu:2017}. The sensory input, as the pole sweeps across a ray at angle $\phi\degree$, increased linearly from 0 at $(\phi-1)\degree$ reaching the maximum value at $\phi\degree$ and falling back to 0 at $(\phi+1)\degree$ and vice verse. Note that the agent sensed the pole only when it intersected a ray but it ``disappeared'' from view while passing between rays. The pole was considered dropped if it fell beyond the rays, or if the agent moved farther than 45 units on either side from its starting position. Performance in this task was calculated by averaging cosine$(\theta*6)$ at each time step of the 500s evaluation duration, where $\theta$ is the pole angle with the vertical axis. To promote generalization of the behavior, performance was averaged over 16 trials with the pole starting from 4 different angles on either side of the agent in $[-9\degree,9\degree]$ with angular velocity -0.1 or 0.1.

\subsection{Evolutionary optimization}
An evolutionary search algorithm was used to optimize the parameters of the agent: time-constant, gain and bias for sensory neurons ($3$), weights from sensory layer to $N$-interneurons ($N*7$), recurrent weights between interneurons ($N^2$), bias and time-constant for each interneuron ($2N$), weights from interneurons to motor neurons ($2N$) and gain, bias and time-constant for motor neurons ($3$): totaling $D = 3+7N+N^2+2N+2N+3$ parameters. A search started with a random population of 100 solutions encoded as $D$-dimensional genotype vectors with each element in [-1,1]. These elements were scaled and mapped on to the different parameters to build the agent. Gains are scaled to be in [1,20], time constants in [1,2], biases in [-4,4] and all weights were scaled to be in [-5,5]. The fitness of agents was evaluated based on their performance in each task. Based on fitness, an elitist fraction of the top 4\% solutions were retained while their copies were subject to a Gaussian mutation noise with mean 0 and variance 0.3 to produce a new population of solutions. This repeated for a fixed number of generations.

Since optimization is stochastic, 100 independent runs were carried out for the single and multi-task scenarios. For the individual tasks, optimization was carried out for 1000 generations in each run. In the multi-task setting, these experiments were conducted in three different task presentation paradigms: (1)~evolved for both categorization and pole-balancing for 2000 generations, (2)~evolved only for pole-balancing for the first 500 generations, and then evolved for both tasks for 1500 generations and (3)~evolved for categorization for the first 1000 generations, and then for both tasks for another 1000. The 500 generation limit for paradigm 2 and 1000 for 3 was based on the number of generations required to acquire good performance in each task when optimized individually. Agents were reset between all trials of all tasks. In the multifunctional cases, the product of the individual task fitnesses was used as opposed to sum or average because it guarantees good performance in both tasks, while still keeping the fitness in $[0,1]$. All three optimization paradigms gave similar results.

\section{Results}
\label{sec:results}
\subsection{Minimal neural resource requirement for each task}
\begin{figure}[!t]
    \centering
    \includegraphics[width=3.2in]{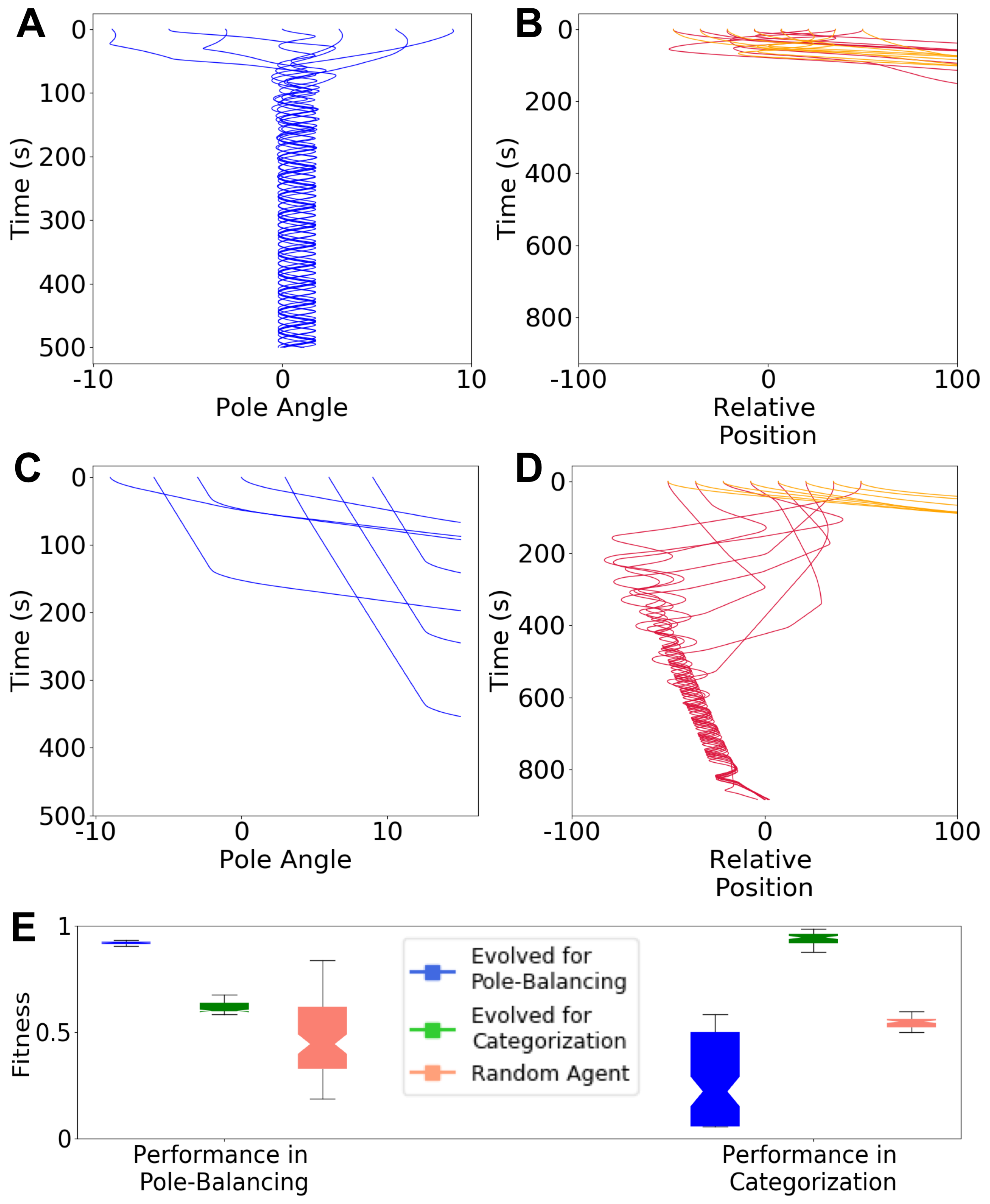}
    \caption{Behavior and performance on individual tasks.
    [A]~Best agent from 100 runs of optimizing for pole-balancing alone. The agent was able to bring the pole to the center and keep it balanced from different initial pole angles.
    [B]~Best pole-balancer shown in A is unable to categorize circles (red) from lines (orange) and avoids both.
    [C]~Best agent from 100 runs evolved for categorization alone is unable to balance the pole.
    [D]~Best categorization agent shown in C, demonstrating its ability to catch circles (red) while avoiding lines (orange).
    [E]~Optimizing for one task results in performance similar to a random agent on the other task. Fitness distribution from 100 runs of agents evolved for pole-balancing (blue) in pole-balancing and categorization, and similarly that of the agents evolved for categorization only (green) in pole-balancing and categorization, and random agents (salmon) on both tasks.}
    \label{fig:bestBehav}
\end{figure}

In order to evaluate the level of reuse in the multifunctional networks, we first systematically explored the minimal resources required to solve each task individually. 100 independent evolutionary runs were performed for networks of different sizes for each task. The smallest network that could perform pole-balancing had 2 interneurons. The best of these agents achieved 98.44\% fitness and was able to move the pole to its upright position from a broad range of initial positions and keep it balanced for an extended duration of time (Fig.~\ref{fig:bestBehav}A). The smallest network that could perform the categorization task also had 2 interneurons. The best of these agents had a fitness of 98.5\% and was able to successfully catch all circles and avoid all lines falling from the full range of starting positions (Fig.~\ref{fig:bestBehav}D).

In order to address multifunctionality using these two tasks, it is important to demonstrate that they indeed require their own set of sensorimotor transformations. In other words, circuits that solve one task, should not be able to solve the other task, and vice versa. To demonstrate this, all agents that were optimized to perform one task were evaluated on the other. Agents that were trained to balance the pole were as good as random agents at the categorization task (Fig.~\ref{fig:bestBehav}B,E). Agents that were trained to categorize could balance the pole only slightly better than random controllers (Fig.~\ref{fig:bestBehav}C,E). This suggests each task requires its own unique set of sensorimotor transformations and that ultimately solving one task does not guarantee good performance in the other.

\subsection{Structural network reuse: Fully overlapping circuits were used to perform both behaviors}
\begin{figure}
    \centering
    \includegraphics[width=3.12in]{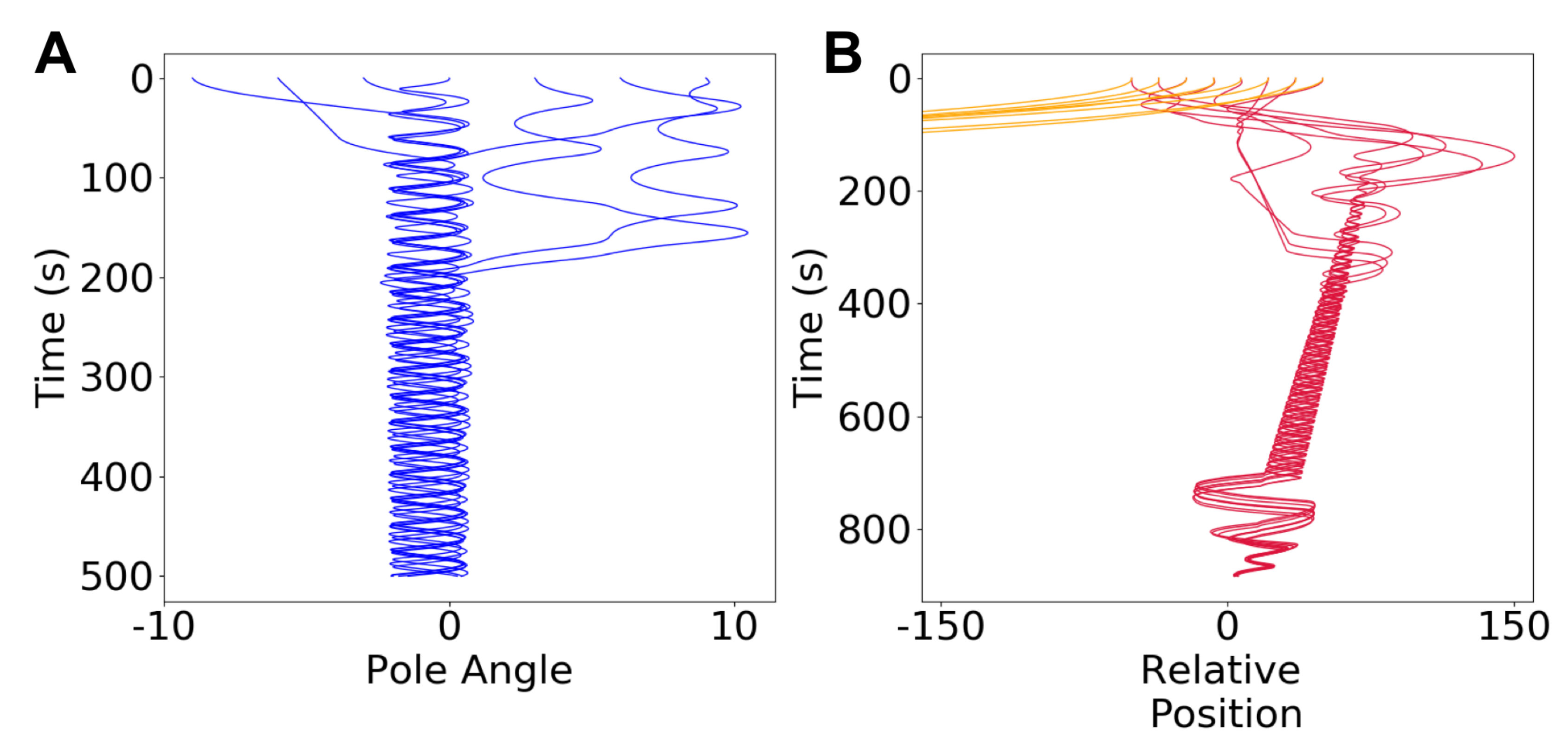}
    \caption{Behavior of the best multifunctional agent from 100 runs.
    [A]~The agent was able to bring to center and balance the pole starting from different pole angles.
    [B]~The same agent, using the same neural network was also able to catch the circles (red) while avoiding lines (orange).}
    \label{fig:multi_behav}
\end{figure}

The highest level of reuse is that of structure - an agent performing more than one behavior could acquire specialized circuitry to perform each behavior or could share neural circuits between them. In order to test this, using the same evolutionary optimization approach described previously we evolved networks of different sizes to perform both behaviors. Interestingly, agents with networks no larger than the ones that could solve the individual tasks could also solve both tasks. The best 2-interneuron multifunctional agent could perform categorization with a fitness of 95.8\% and pole-balancing with a fitness of 95.4\%. The optimization scheme that led to this agent was composed of evolving for pole-balancing for the first 500 generations, followed by evolving for both tasks. This agent used the same circuit to successfully catch circles while avoiding lines and also balance a pole (Fig.~\ref{fig:multi_behav}). It is to be noted that the agent had no external signal indicating which task to solve. Furthermore, the agent also had no synaptic plasticity or neuromodulatory signals that could be responsible for re-configuring the circuit for the different tasks. Since performing each task individually required at least 2 interneurons, it follows that the multifunctional agent used fully-overlapping structural networks to perform both.

\subsection{Autonomous dynamics reuse: Overlapping sets of attractors were used to perform both behaviors}
Given that the circuit is the same across the two behaviors (object categorization and pole-balancing), we wanted to evaluate if there was reuse in the autonomous dynamics of the neural network of this agent. In order to do this, we first constructed the phase-portraits for several inputs across each task. A phase-portrait for a particular input can have one or more attractors. The set of phase-portraits associated with a behavior (say circle catching) can be obtained by fixing the inputs to what the agent experiences during that behavior (circle at fixed positions relative to the agent), and allowing the network to settle into its attractors from different initial states. The three sets of phase-portraits corresponding to circle catching, line avoiding and pole-balancing were compared based on attractor composition, basins of attraction, and location of attractors to evaluate reuse.

Attractor compositions refers to the type of attractors that were present in the set of phase-portrait for each behavior (i.e. fixed-points, limit-cycles etc.). In this agent, all phase-portraits associated with all behaviors in the best multifunctional agent were only composed of fixed-point attractors. Even though attractor composition is the same across behaviors, they could have different basins of attractions around those attractors. This could lead to different behaviors operating in its own region of the phase-space. However, for this agent, since only one fixed-point attractor existed in all phase-portraits, there exists only one basin of attraction which is the same across all behaviors. Thus, with same attractor composition and basins of attraction, the phase-portraits were qualitatively similar across all behaviors (i.e. no bifurcation).

\begin{figure}[t]
    \centering
    \includegraphics[width=1.6in]{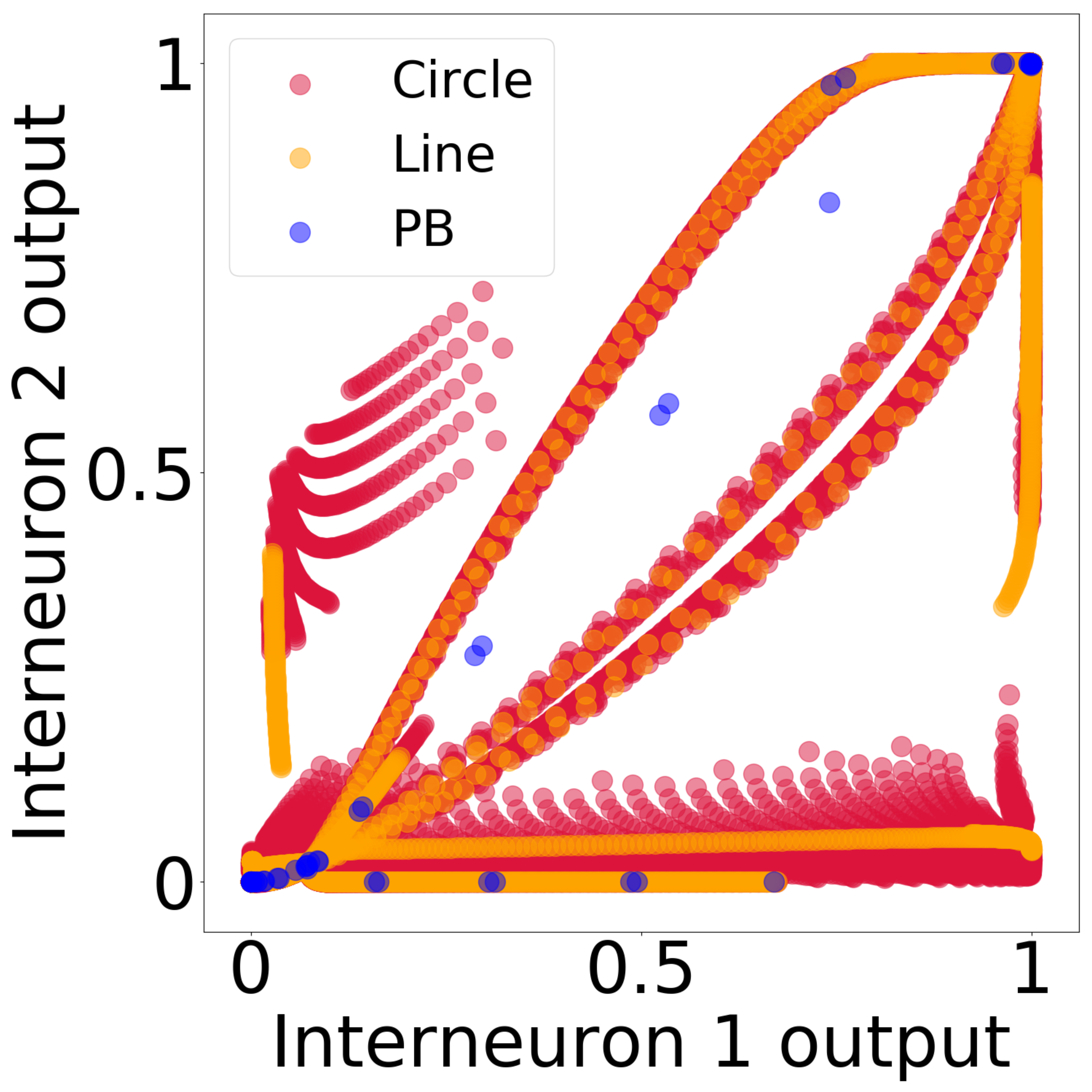}
    \caption{Attractor reuse. Locations of attractors from the three sets of phase-portraits corresponding to circle-catching (red), line-avoiding (orange) and pole-balancing (blue) tasks are overlapped. This shows that each behavior has its own set of unique attractors as well as shares them with other behaviors.}
    \label{fig:att_reuse}
\end{figure}

These qualitatively similar phase-portraits could further be quantitatively compared based on the location of attractors in them. Differentiated by their location, each behavior could have a unique set of attractors or they could overlap to different extents; the exact locations of the fixed-point attractors on all phase-portraits of different behaviors do not have to be the same. Upon analyzing their locations we discovered that the multifunctional agent reused attractors identical in location between these behaviors (Fig.~\ref{fig:att_reuse}). This reuse was only partial since each behavior also had its own set of unique attractor locations that were not shared. Reusing the same attractors means that different inputs from different behaviors were mapped to the same phase-portrait, which suggests that there is an inherent degeneracy between the sensory inputs and the requisite behavioral pattern.

\subsection{Transient dynamics reuse: Phases of the behaviors reused the same transient dynamics}
Our analysis started at the level of structural reuse and went on to discover reuse at the level of autonomous dynamics in the best multifunctional agent. The next level is that of ongoing dynamics as the nervous system coupled with its body and environment performs the behaviors. Note that in the previous level, attractors were identified by fixing the relative position of the agent with the object and then allowing the dynamics of the network to settle to their attractors. However, during behavior, both the agent and the object are in constant movement. Therefore, at any given time, for a particular relative positioning of the agent and the object, the sensory input might change before the network settles into the attractor associated with that fixed input. As a result, dynamics of the network are in constant transient movement across the phase-portraits (and the attractors therein) associated with that behavior. Since attractors in the same location were partially reused in this agent, the relevant question to explore reuse at the next level is whether multiple behaviors have unique transients or if they could be shared.

\begin{figure}[t]
    \centering
    \includegraphics[width=3.5in]{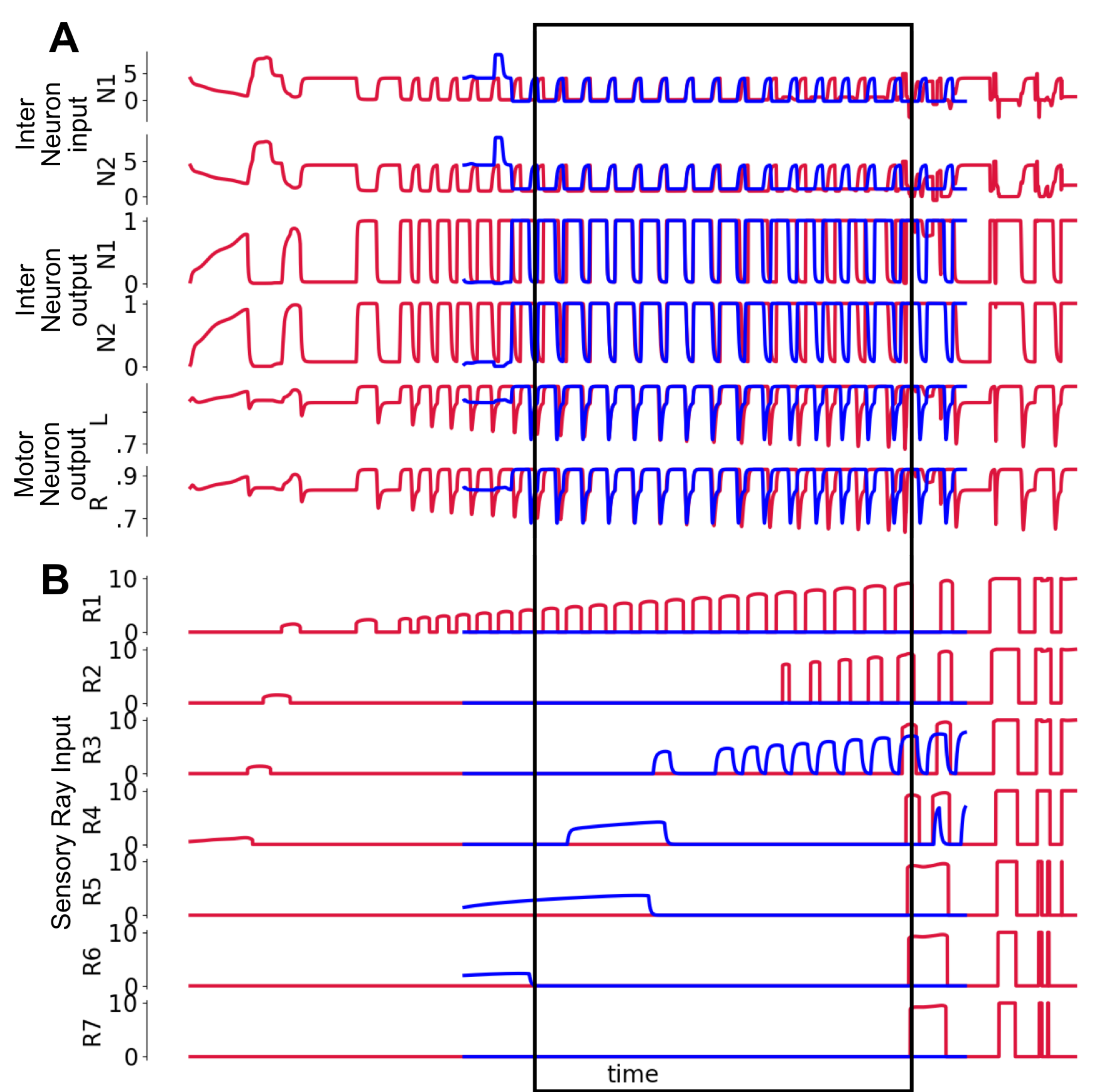}
    \caption{Transient/Driven dynamics reuse.
    [A]~Activity of inter-neurons for the circle-catching (red) and pole-balancing (blue) tasks, time-shifted to show identical neural activity.
    [B]~Sensory inputs to the 7 rays showing that although neural activity is indistinguishable, the agent tracks the circle along ray 1(red) but the pole is along rays 3 and 4 (blue).}
    \label{fig:transientReuse}
\end{figure}

Transient dynamics were shared partially between the circle-catching and pole-balancing tasks in the best multifunctional agent. In order to understand the behavioral implications of this, we evaluated the entire sensorimotor loop as this agent performed both behaviors. While in each case there were times when their transients were different, for a particular phase during these behaviors, the dynamics almost exactly matched (Fig.~\ref{fig:transientReuse}A). Inputs to the interneurons, their outputs, and motor neuron outputs were all identical. This suggests that the agent's nervous system does not differentiate between these phases of the two behaviors. This leads to two interesting questions: (1) Are the two behaviors indistinguishable during this phase? (2) If not, where does the difference come from?

Although the transient dynamics in the inter-neurons were identical, the behaviors were different in the circle-catching and pole-balancing tasks. In the former, the agent received oscillatory inputs only along ray 1, meaning that the agent oriented itself so as to track the circle along ray 1 before catching it (Fig.~\ref{fig:transientReuse}B). In contrast, the sensory inputs during the same transient dynamics in the pole-balancer shows that the agent maintains the pole oscillating around ray 3 bringing it across 4,5 and 6. This is an interesting outcome demonstrating that neural activity that is completely indistinguishable can still produce behaviors starkly different from each other. The difference arises from the parts of the behavior leading up to this shared transient phase, where the agent has its own unique dynamics for each behavior through its interaction with the environment. Note that the weights to the interneurons from sensory rays 1 and 3 are not the same. Thus, transient reuse emerges purely from brain-body-environment interaction.

\section{Related work}
The work presented here builds on previous work using brain-body-environment computational models for multiple tasks \cite{Izquierdo:2008, Williams:2013, Agmon:2014} by developing a computational model of a brain-body-environment system that performs multiple behaviors using the same sensory and motor capacities. In the work by Izquierdo et. al. (2008), the same neural network without any changes in parameters was shown to perform two qualitatively different behaviors while placed in two different bodies. Williams et. al. (2013) showed that when different motor systems are used for different tasks, the qualitative difference in environmental feedback drives the same network differently to produce different behaviors. Agmon et. al. (2014) presented a model where different sensory apparatus in the agent, sensitive to different stimuli, performed different associated behaviors using the same motor control systems. In these models, although the neural network remained the same, the body was changing. The model presented in this paper, used the same sensory and motor control mechanisms for the two tasks - object categorization and pole-balancing. We also show through dynamical analysis that reuse to the level of transient dynamics can be observed when the brain, body, environment and their interaction are taken into account.

\section{Discussion}
\label{sec:discussion}
To summarize, we first evolved embodied recurrent neural networks to perform object categorization, pole-balancing and then both. We then systematically explored the different levels of neural reuse in the evolved circuits. We discovered reuse of neural circuits at the structural level, followed by reuse of autonomous dynamics with qualitative sharing of phase-portraits, overlapping basins of attraction and reuse of attractors identical in the location of their fixed-points. Furthermore, we discovered partial reuse of transient dynamics in the best multifunctional agent. The two main contributions of our work are as follows: (1) the same neural circuit can perform multiple behaviors using the same sensory and motor systems in the absence of explicit task identifying signals or processes such as neuromodulation; (2) indistinguishable neural activity, displaying reuse to the level of transient dynamics, can still produce completely different behaviors.

The rationale behind transient reuse in this multifunctional agent can be explained by analyzing the environment-body relationship and transient dynamics. The similarity in dynamics arises out of the agent's ability to generalize between the two behaviors by learning to align an object along a single ray - pole along center ray and circle along corner ray (Fig.~\ref{fig:multi_behav} and~\ref{fig:transientReuse}). Generalization requires learning to use only one ray because the pole only intersects one ray at a time. The circle had to be balanced along the corner ray because otherwise it would intersect multiple rays and the pole needs to be balanced along the center ray to maximize fitness. The difference in behavior, however, arises out of the unique transient dynamics prior to shared transient phase of the behaviors. The unique dynamics in circle-catching orients the falling object along the corner ray even if the circle starts from the center, whereas in pole-balancing it brings the pole to the center, thereby setting up the system to perform generalized object tracking along a single ray for both behaviors (Fig.~\ref{fig:multi_behav}B). This is possible because of the structure provided by the environment and the body. Objects intersect only one ray or multiple rays, yet the agent is required to align with the object in both cases. Multifunctionality in this agent is made possible by the closed-loop interaction between brain, body, and environment -- a possibility that is not typically taken into consideration in the literature \cite{Briggman:2008}. Therefore, our results expand the list of possible mechanisms that enable multifunctionality in living organisms.

Due to experimental limitations, the study of multifunctionality has been mostly concerned with motor neuron circuits capable of generating multiple patterns of activity~\cite{Briggman:2008}. Here we extend this framework to circuits that are behaviorally multifunctional: from sensory input, through interneurons, to motor neurons responsible for generating actions. We demonstrate that multifunctionality can result from the closed-loop interaction between brain, body, and environment. Therefore, our results expand the list of mechanisms that can result in multifunctionality to include closed-loop interactions. Ultimately, this mechanism can coexist with previously described mechanisms, including  neuromodulation and synaptic plasticity.

The three-level framework presented is rooted in dynamical systems theory and has the potential to explain neural reuse in any behavioral system, biological or artificial. We intentionally focused on a small neural controller and a simple set of behavioral tasks. However, the possibilities uncovered in this system should be available to larger neural networks solving more complicated tasks. Ultimately, embodied neural reuse offers a distinct perspective on several topics of interest to understanding cognition, including modularity in brain organization, localization of cognitive functions, and more generally the mapping between brain structure and function.

\section{Acknowledgments}
The work in this paper was supported in part by NSF grant IIS-1524647.

\bibliographystyle{apacite}

\setlength{\bibleftmargin}{.125in}
\setlength{\bibindent}{-\bibleftmargin}

\bibliography{refs}

\end{document}